# CapsNet comparative performance evaluation for image classification

Rinat Mukhometzianov[1] and Juan Carrillo[1]

[1] University of Waterloo, ON, Canada

**Abstract.** Image classification has become one of the main tasks in the field of computer vision technologies. In this context, a recent algorithm called CapsNet that implements an approach based on activity vectors and dynamic routing between capsules may overcome some of the limitations of the current state of the art artificial neural networks (ANN) classifiers, such as convolutional neural networks (CNN). In this paper, we evaluated the performance of the CapsNet algorithm in comparison with three well-known classifiers (Fisherfaces, LeNet, and ResNet). We tested the classification accuracy on four datasets with a different number of instances and classes, including images of faces, traffic signs, and everyday objects. The evaluation results show that even for simple architectures, training the CapsNet algorithm requires significant computational resources and its classification performance falls below the average accuracy values of the other three classifiers. However, we argue that CapsNet seems to be a promising new technique for image classification, and further experiments using more robust computation resources and refined CapsNet architectures may produce better outcomes.

## 1 Introduction

Classification predictive modeling is the task of receiving a new observed sample (X) as input and assigning it to one of the predefined categories (y), also known as labels, by using a trained model (f). This classification task creates the basis for other computer vision problems such as detection, localization, and segmentation [1]. Despite the fact that this task can be considered as straightforward for humans, it is much more challenging for a computer-based system; some of the complications are viewpoint-dependent object variability and the high in-class variability of having many object types [2].

Today, engineers and researchers around the world use convolutional neural networks (CNNs) to solve common problems in the field of image classification, this technique has produced remarkable low test errors on classification tasks over different types of images [3,4]. Despite their success, CNNs also have several drawbacks and limitations. CNNs accumulate sets of features at each subsequent layer; it starts from finding edges, shapes, and finally actual objects. However, most of the infor-mation about the spatial relationships (perspective, size, orientation) between these features is lost. They seem to be easily fooled by images with features in the wrong place (for instance a nose instead of an eye on a human face), or samples of the same images but in different orientations. One way to eliminate this problem is with excessive training for all of the possible angles, but it usually takes a lot more time and computational resources [5]. Moreover, convolutional neural networks can be susceptible to white box adversarial attacks [6] and the so-called "fast gradient sign method" [7]. A new algorithm called dynamic routing between capsules (CapsNet) was recently proposed to overcome these drawbacks [8].

The idea of translated replicas of learned feature detectors forms the basis of CNN's. In other words, information about properly trained features gathered in one position can be spread out to other positions; this functionality has become advantageous for image interpretation. In contrast, CapsNet replaces the scalar-output feature detectors from CNNs with vector-outputs, it also replaces the max-pooling subsampling technique with routing-by-agreement, so it enables the duplication of learned knowledge across space. In this new CapsNet architecture, only the first layer of capsules, also known as primary capsules, includes groups of convolutional layers. Following traditional CNN ideas, higher-level capsules cover more extensive regions of the image. However, the information about the precise position of the entity within the region is preserved in contrary to standard CNNs.

For lower-level capsules, the location information is "place-coded" by the active capsule. As we ascend in the hierarchy, more and more of the positional information is "rate-coded" in the real-valued components of the output vector of a capsule. All these ideas imply that the dimensionality of capsules must increase as we move up in the hierarchy. To summarize, the Capsule Networks give us the opportunity to take full advantage of intrinsic spatial relationships and model the ability to understand the changes in the image, and thus to better generalize what is perceived.

## 2 Datasets

The Yale face database B (Figure 1) contains 5850 grayscale images distributed in 38 classes, with varying illumination conditions and points of view [9]. The dataset is a good candidate to evaluate the performance of capsule networks and their capabilities to model object features from different points of view.

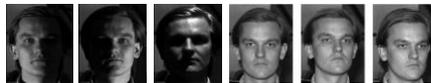

**Fig. 1.** Example images from the Yale face database B.

The MIT CBCL dataset (Figure 2) includes 5240 images with different conditions of illumination and pose [10]. The characteristics of the images in this dataset are similar to those of the Yale face database B.

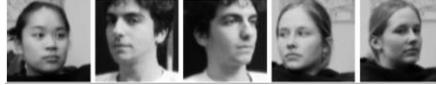

**Fig. 2.** Example images from the MIT CBCL dataset.

The Belgium TS dataset (Figure 3) comprises 7000 images of 62 types of traffic signs from urban areas of Belgium [11]. Some data collection aspects such as the illumination level, point of view, and distance, difficult the classification process. Additionally, several limitations including dirt, stickers, trees, and typical deterioration produce an occlusion effect in which only some parts of the traffic signal were recognizable.

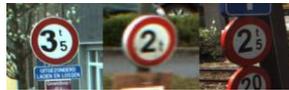

**Fig. 3.** Example images from the Belgium TS dataset.

The CIFAR-100 dataset (Figure 4) includes 60,000 images of 100 types of everyday objects, such as vehicles and animals [12]. From the four datasets, this is the one that represents the biggest challenge for an image classification task, mainly due to a significant variation (illumination, pose, size, etc.) between images from the same class.

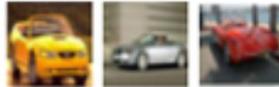

**Fig. 4.** Example images from the Cifar-100 dataset

## 3 Literature review

### 3.1 Baseline methods

Since the main goal of our project is to evaluate the performance of CapsNet on different types of images (faces, traffic signs, everyday objects) we compared its performance with other suitable methods highlighted in the literature as reliable and robust for every type of datasets. First, we compare CapsNet with the Fisherface algorithm [13] for face datasets (Yale face database B and MIT). This algorithm has proved to be fast, reliable [14,15], and one the most successful methods for face recognition [16,17] achieving an average of 96.4% recognition rate on the Yale B Extended Database [18] and 93.29% on the MIT CBCL dataset.

For the Belgium Traffic signs dataset, we chose a CNN architecture known as Le-Net-5 [19] and designed initially for handwritten digit recognition. The advantage of having a CNN is that it requires a much lower number of trainable parameters as compared to a Multi-Layer Feed Forward Neural Network, which supports sharing of weights and partially connected layers. Along with a reduced number of trainable parameters, a CNNs design also makes the model invariant to translation, a distinctive feature of state-of-the-art methods for image classification. LeNet-5 is relatively small but has proved to be powerful enough to solve several classification problems including traffic sign recognition [20].

A recent implementation of a LeNet classifier achieved 95% accuracy on the German traffic dataset [21]. Although we did not find any studies implementing LeNet on the Belgium TS dataset, a similar implementation of a simple Fully Connected network with two layers achieved 70% of accuracy [22]. Everyday objects classification is still a challenging computer vision problem. One of the state-of-the-art methods for this task is called Deep Residual Learning for Image Recognition (ResNet) [23]. It was presented in configurations with 18, 34, 50, 101, 152 layers. The best top-5 error achieved on ImageNet test dataset (1000 classes of different objects naturally appearing in the world) [24] is 3.57% with an ensemble of six ResNet.

### 3.2 Capsule networks

The article that introduces the CapsNet architecture is called *Dynamic Routing Between Capsules* [8]. In this article, a novel type of neural network model for image classification is provided. The main advantage is that this model preserve hierarchical spatial relationships; theoretically, this architecture may learn faster and use fewer samples per class. The article describes an end-to-end trainable model of CapsNet and provides modeling results. On the MNIST dataset, 0.25% classification error rate was achieved, 5.2% on MultiMNIST, 10.6% on CIFAR10, 2.7% on smallNORB.

Another recent study implements the CapsNet design and explores different effects of model variations, ranging from stacking more capsule layers to changing hyperparameters [25]. Also, a new activation function was presented. They used fewer epochs due to computational constraints and finished with 2.57% accuracy improvement over the baseline MNIST model introduced in the previous paper. In the third related study that we found [26] a new type of convolutional architecture called Non-Parametric Transformation Networks (NPTNs) was introduced. In this model, Capsule Nets are augmented with NPTNs, replacing the convolution layers in the Primary Capsule layer with NPTN layers, while maintaining the same number of parameters. They gained 1.03% improvement in test error on MNIST dataset.

## 4 Task and methods

This project focus on evaluating a new artificial neural network (ANN) architecture called CapsNet [8] to addresses the problem of image classification. According to the authors, this algorithm may find practical applications in deep learning because these capsules can better preserve and model hierarchical relationships inside of the internal knowledge representation of a neural network. Our goal is to implement the CapsNet architecture, test it on multiple datasets, and compare the experimental results with other high-performance classification methods mentioned in the literature review section.

Fisherfaces is an example of a classic image classification method, in the sense that it tries to "shape" the scatter to make it more reliable for classification. This method selects W in such a way that the ratio of the between-class scatter and the within-class scatter is maximize. W is the matrix of a linear transformation of input images to the new feature space. It has orthonormal columns, and the tuning parameter is the number of Fisherfaces to keep.

LeNet-5 is a Convolutional Neural Network, it includes convolutional layers, average pooling layers and Fully Connected (FC) layers. Also, activations nonlinear function such as sigmoid and Tanh. Like most of the standard convolution networks with fully connected layers, we can tune several parameters of the architecture, such as the activation function, number of hidden layers and units, and weight initialization method. Also, specifically for this CNN architecture, the window size, stride value, and some filters at each layer.

ResNet is the type of architecture, in which the main building component is a residual block, which is a stack of regular layers such as convolution and batch normalization, but its input flows not only through the weight layers but also through the identity mapping shortcut. Finally, as can be seen in Figure 5, the two paths are summed up. Resnet gives us the opportunity to stack as many residual blocks as necessary, the standard number of layers proposed in [23] are 18, 34, 50, 101, and 152.

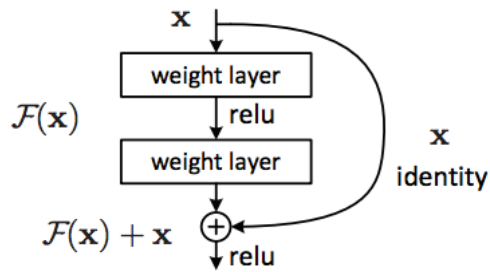

**Fig. 5.** A residual block, the fundamental building block of the residual networks

Although the CapsNet architecture was briefly presented in the Introduction, we will review it here in more detail. A capsule network (CapsNet) is made of a network of capsules; a capsule is a small group of neurons each referring to a particular component. The activity of neurons in a group determine the properties of the component in the image. Each capsule is responsible for identifying a single component. All the capsules together do what a traditional CNN does, assigning the correct label to an image. Unlike traditional CNNs, CapsNet preserves the location and orientation of each component within an image.

The CapsNet algorithm is built with multiple numbers of layers, similar to conventional neural networks. Each one of the low-level capsules, also known as primary capsules, receives a small region of the image as input (called its receptive field) and tries to detect the presence and pose of a particular item, for example, a rectangle. Capsules in higher layers called routing capsules, detect larger and more complex objects.

In CapsNet, the output of each capsule is a vector, not a scalar, so we cannot use, for example, the standard softmax or sigmoid activation function. In contrast, the non-linear "squashing" function (Equation 1) is introduced to ensure that short vectors get shrunk to almost zero length, and long vectors get shrunk to a length slightly below 1. We denote $s_j$ as the input of a capsule, and $v_j$ as its output vector.

$$v_j = \frac{\|s_j\|^2}{1 + \|s_j\|^2} \frac{s_j}{\|s_j\|^2} \quad (1)$$

This innovative concept is called routing by agreement. In this process, the information from capsules in the primary layer is only supplied to the upper capsules layer if the detected item has been successfully handled by those two capsules in the past. Sabour et al. affirm that this type of "routing-by-agreement" is far more effective than the primitive form of routing implemented by max-pooling.

CapsNet exploits the length of the instantiation vector to represent the probability that a capsule's entity exists. Routing capsules on the top level will see a long instantiation vector only if the object is present on the image. A separate margin loss was proposed to allow for multiple categories (Equation 2).

$$L_k = T_k \max(0, m^+ - \|v_k\|)^2 + \lambda (1 - T_k) \max(0, \|v_k\| - m^-)^2 \quad (2)$$

Where $T_k$ equals 1 if a digit of class k is present and $m^+ = 0.9$ and $m^- = 0.1$. The $\lambda$ down-weighting of the loss is set to 0.5 by default.

They also suggest using an additional reconstruction loss to encourage the routing capsules to encode the instantiation parameters of the input. During training, they mask all but the activity vector of the correct routing capsule. Then, the algorithm uses this activity vector to reconstruct the input image. The output of the routing capsule goes into a decoder consisting of 3 fully connected layers, a mechanism to minimize the sum of squared differences between the outputs of the logistic units and the pixel intensities. As we can see, the CapsNet algorithm includes more parameters in comparison with a standard neural network, such as the number of dimensions in each capsule type (primary or routing capsules), the number of primary and routing capsules, and the number of channels in capsule layer.

## 5    Implementation

### 5.1    Software tools

Python 3.6 was chosen as a programming language and Spyder IDE as a main programming environment. The main learning frameworks we used were Tensorflow 1.6 [27], Keras 2.1.5 [28], and OpenCV 3.4.1 [29]. For data preprocessing we use Numpy and Scipy libraries [30]. Regarding data structures, Tensorflow and Keras perform all computation on Tensors, while Numpy and Scipy use arrays. Moreover, Python set and list structures were used to store labels. All four datasets are represented by 2D gray-scaled images and their corresponding class labels.

### 5.2 Computing resources

For running the classification algorithms, we used a Dell laptop and some virtual instances hosted on the Google Cloud Platform. The Dell laptop has one Nvidia Ge-Force 1050 Ti Mobile GPU, 16Gb of RAM, and an Intel Core i7 extreme 7th generation CPU. The available configuration of instances in Google Cloud includes one Nvidia Tesla K80, 5Gb of RAM, and an Intel Xeon (2.0 GHz) CPU.

### 5.3 Preprocessing

Several preprocessing techniques were applied to the datasets, depending on their characteristics and the requirements of the CapsNet algorithm. Table 1 summarizes the preprocessing techniques applied to each dataset. In our project, our goal was to evaluate and compare the performance of the CapsNet algorithm using the same number of images included in the original datasets, so we did not use any augmentation techniques.

**Table 1.** Preprocessing per dataset

| Dataset | Grayscale color space | Min-max norm. [0,1] | Histogram equalization | Resize (width×height) |
|---------|----------------------|---------------------|------------------------|------------------------|
| Yale face database B | Originally in grayscale | Applied | Applied only to some images | From 192×168 px. to 96×84 px. |
| MIT CBCL | Originally in grayscale | Applied | Applied only to some images | From 200×200 px. to 72×55 px. |
| Belgium TS | From RGB to grayscale | Applied | Not applied | From different sizes to 90×90 px. |
| CIFAR-100 | From RGB to grayscale | Applied | Not applied | Not applied, original size 32×32 px. |

### 5.4 Training architectures

Training, validation and testing sets were obtained as 70%:15%:15% from every dataset. We utilized Adam as a state-of-the-art optimizer [31]. For the two face datasets we use the standard Fisherfaces algorithm, which works in two steps: First, it generates Fisherfaces, project probe, and reference faces (or images) into a dimension-reduced subspace, then it calculates Euclidean distances between reproduced probe faces and reference faces and get the minimum distance value candidate.

For the Belgium TS, we used a modified version of LeNet-5. The standard version of LeNet-5 receives input images of 32x32 pixels; but to preserve more information, we only resized the images down to 90x90 pixels and added more convolutional layers. We found the optimal learning rate is 0.0001 using twenty K-fold (K=5) validation split tests. Batch size for the modified LeNet was set to 128. To summarize, the overall structure is shown in Table 2.

**Table 2.** Modified LeNet-5. Padding is valid for the convolutional layer, ReLu is the activation function for all layers.

| Layer | Type | Input | Kernel size/stride | Output |
|-------|------|-------|-------------------|--------|
| 1 | Convolutional | 90x90x1 | 7x7 | 84x84x6 |
|   | Pooling | 84x84x6 | 2 | 42x42x6 |
| 2 | Convolutional | 42x42x6 | 7x7 | 36x36x16 |
|   | Pooling | 36x36x16 | 2 | 18x18x16 |
| 3 | Convolutional | 18x18x16 | 7x7 | 12x12x32 |
|   | Pooling | 12x12x32 | 2 | 6x6x32 |
| 4 | Flattening | 6x6x32 |  | 1152 |
| 5 | Fully Connected | 1152 |  | 300 |
| 6 | Fully Connected | 300 |  | 200 |
| 7 | Fully Connected | 200 |  | 62 |

In the CIFAR-100 dataset, we selected ResNet as the baseline algorithm. Training a ResNet network with 50 layers may take up to 29 hours using eight Tesla P100 GPUs; therefore, we applied a transfer learning technique [32]. Moreover, we selected a ResNet model (50 layers, 92.9% top-5 accuracy) which was pre-trained on the ImageNet dataset and retrained only the last five layers, setting the learning rate as 0.000001. The batch size for the ResNet-50 transfer learning model was set to 64.

To implement the CapsNet model, the first layer is a standard convolutional, we took this design from Sabour et al. [8] and used it in the same way for all our tests.

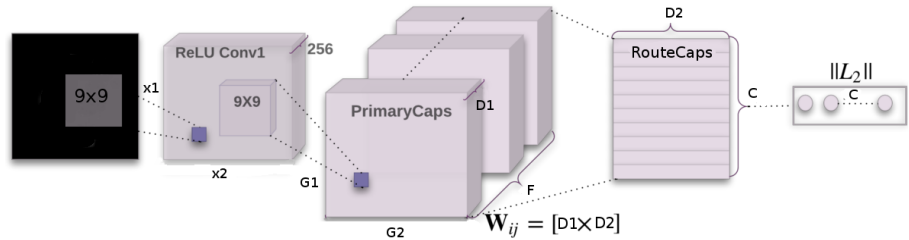

**Fig. 6.** CapsNet architecture.

This first convolutional layer produces 256 feature maps, using a 9x9 kernel and valid padding. For the convolution within the primary capsules we also kept a kernel size of 9x9, x1 and x2 entirely depend on the size of the input image. G1xG2 are the dimensions of each capsule, and these values are computed automatically based on x1 and x2. Other parameters we tuned for different datasets including D1 and D2 (the dimensions of the output vectors in primary and routing capsules), F (the number of channels in the primary capsule layer), and C (the number of classes).

To test different hyperparameter options we tested several batch sizes for CapsNet, from 10 to 50 to fit GPU RAM. The end-to-end CapsNet architecture is depicted in Figure 6.

# 6 Results and evaluation

The performance of the CapsNet algorithm was evaluated on four datasets, which were selected because they have been widely used for evaluation of image classification algorithms and their results have been published in renowned journals or conference proceedings. As a metric for comparison, we selected classification accuracy. The overall classification results are listed in Table 3.

**Table 3.** Comparison of classification results.

| | | | | Baselines | | CapsNet | |
|---|---|---|---|---|---|---|---|
| **Dataset** | **Classes** | **Instances** | **Algorithm** | **Avg. training time** | **Test accuracy** | **Avg. training time** | **Test accuracy** |
| Yale Face Database B | 38 | 5850 | Fisherface | ~5 minutes* | 98.2%** | ~24 hours*** | 95.3% |
| MIT CBCL (faces) | 10 | 5240 | Fisherface | ~1 minute* | 98.3%** | ~14 hours*** | 99.87% |
| BelgiumTS (traffic signs) | 62 | 7000 | Modified LeNet | <1minute* | 98.2% | 16 hours*** | 92% (40 epochs) |
| CIFAR-100 (objects) | 100 | 60000 | Resnet 50 | 20 hours (200 epochs) | 65.5% | 18 hours*** | 18% (35 epochs) |

*Core i7 extreme 7th generation, 16 GB RAM, Nvidia GeForce 1050 Ti Mobile*

**With best parameters we discovered using multiple tests with training and validation sets*

***Google cloud, Intel Xeon, 5 Gb RAM, NVIDIA Tesla K80*

## 6.1 Faces datasets

According to our literature review, the Fisherfaces classifier has reached an accuracy of 94.5% on average for both datasets including face images (Yale and MIT). In our implementation of Fisherfaces, we achieved a slightly higher accuracy in both face image datasets, likely due to the preprocessing we conducted to enhance image contrast. Additionally, we performed ten 5-fold validations on Fisherfaces of MIT and Yale datasets and made an interesting finding; only 40 Fisherfaces were necessary to obtain the resulting accuracies.

On the other hand, in our implementation of the CapsNet algorithm (D1=8, F=8, D2=8) over the Yale dataset the classification accuracy was quite similar (95.3%) to the average results from Fisherfaces papers and lower than our Fisherfaces results. We consider that CapsNet may have better results with more training epochs; however, due to time constraints and limited resoruces avalaliable in our project, we were not able to extend the training time to get better results (Figure 7). On the MIT dataset, CapsNet (D1=8, F=32, D2=16) outperformed Fisherfaces on the test set; however, the CapsNet required training time exceeded by far that of the Fisherfaces algorithm.

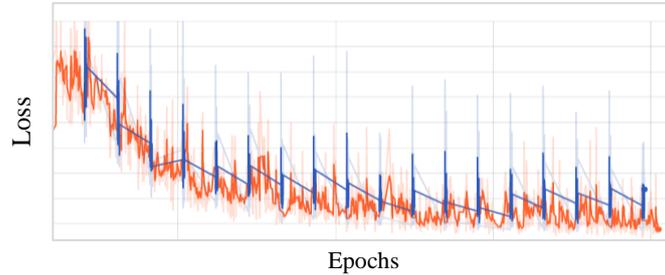

**Fig. 7.** CapsNet algorithm. Training (orange) and validation (blue) loss for the Yale dataset.

### 6.2    Belgium TS

Although no previous studies were found about the use of LeNet on the Belgium dataset, we use the classification accuracy (95%) achieved by Jacopo Credi on the German dataset as a reference to evaluate our results. We obtained a 98% accuracy with the modified version of LeNet. In contrast, after training the CapsNet algorithm for 16 hours (40 epochs), we scored an accuracy of 92%, an outcome that confirms CapsNet as an accurate but less efficient classifier, at least in our experimental setting. We argue that further tuning of hyperparameters might improve the performance.

### 6.3    CIFAR-100

ResNet it today one of the best image classification methods, despite the significant computing requirements that it demands. After 20 hours of training (200 epochs), the resulting accuracy of ResNet-50 is 65.5%, and that is higher than the accuracy obtained from the CapsNet (D1=8, F=8, D2=8) classifier (18%), which only completed 35 epochs after 18 hours of training (Figure 8).

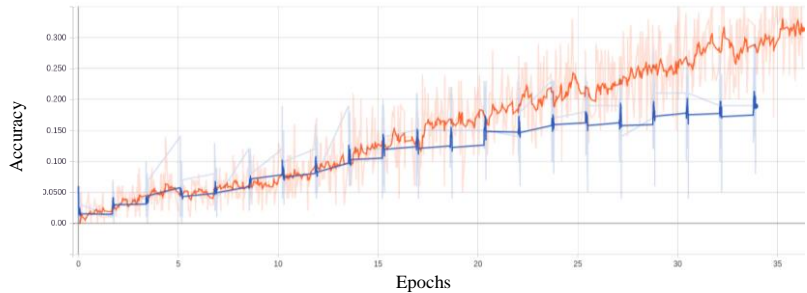

**Fig. 8.** CapsNet algorithm. Accuracy on the CIFAR-100 dataset.

Unfortunately we have limited time and resources for our project, so based on Figure 8 and Figure 9, we suppose to think that given more time the loss function may keep decreasing; however, a 100 classes dataset seems to be a very complex problem for such a 'small' network. Using higher values for D1, F, and D2, the CapsNet model might produce better results.

We also trained CapsNet with two other combinations of parameters (D1=4, F=8, D2=16) (D1=8, F=16, D2=16), but with these settings, the loss function stopped decreasing after around ten iterations, and the accuracy did not grow more than 10%.

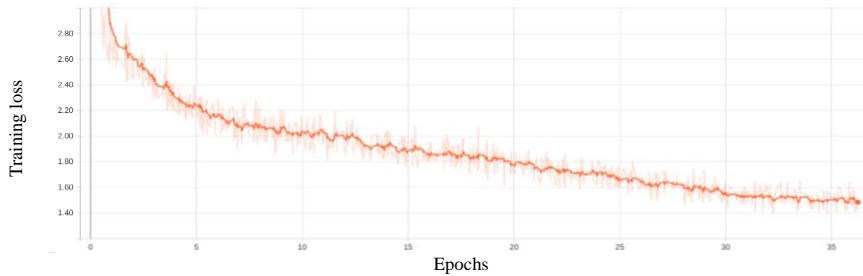

**Fig. 9.** CapsNet algorithm. Training loss on the CIFAR-100 dataset.

## 7    Discussion of results and conclusions

We found that even with a small number of hyperparameters, CapsNet can efficiently classify faces (on a given datasets). Additionally, we found that CapsNet, like classical neural networks, requires more samples per class to achieve a lower test error. Good results on baseline and CapsNet architectures for face datasets also may be result of small variation within classes (relatively small degree of horizontal and vertical face rotation, relatively fixed face position with respect to image).

Regarding training times, we assume that it might take weeks (or months) to achieve a good accuracy on 100 classes with 5000 images per class (CIFAR-100) using a Tesla K80 GPU. Training CapsNets requires more computational resources than ConvNets because the outputs of primary capsules are activity vectors (with instantiation parameters) rather than scalars, leading to higher memory requirements because of the increased dimensionality. Moreover, when increasing the size (height px., width px.) of input images, the number of cells used in GPU RAM increases near exponentially.

We argue that CapsNet has the potential to achieve a higher performance with some modifications in the hyperparameters. Having an 8D vector for routing capsules seems not enough for a complex dataset such as CIFAR-100, and 16D, 32D, 64D, may give a boost in performance, but we need more powerful GPU(s) to fit such complex model to memory. Another alternative could be decreasing the batch size to 1 sample, which may be harmful to the training process because the gradient might go in a wrong direction. Having good preprocessed dataset may solve this problem because in that case, every new sample pushes the gradient on the right track, but it may take longer to converge [33].

In this project, we evaluated the performance of the CapsNet algorithm on four datasets containing different object types such as faces, traffic signs, and everyday ob-jects. Compared to the previous studies found in the literature where the CapsNet algorithm was applied to image datasets of only ten classes [8,25,26] we applied the CapsNet algorithm to datasets of up to 100 classes.

Classical machine learning methods and neural networks adequately designed and tuned can still outperform capsules when we consider more than ten classes. Despite the claims from previous studies suggesting that CapsNet can be trained with a smaller number of samples in comparison with classical CNNs, we found that more complex image scenes still require many images as classical CNNs. A current limitation is that published papers about CapsNet are not very detailed; hence, some open questions about specific aspects of the network implementation are still unanswered, mainly because the authors do not provide thorough recommendations on how to design capsule networks. Increasing dimensions of input images lead to exponential growth of required resources; consequently, running the algorithm over images in their original size becomes very resource demanding. An alternative is to downsize the images to ⅓ or ⅔ of their original size; however, it causes information loss, which consequently diminishes the classification accuracy.

The current state of CapsNet research is today on the same level of advancement as CNNs were in 1998; therefore, more research, experimentation, and tests are required to revealing the full potential of this method.